\documentclass[runningheads]{llncs}
\usepackage{cite}
\usepackage{amsmath}
\usepackage[T1]{fontenc}
\usepackage{graphicx}

\begin{document}

\title{Transforming Multimodal Models into Action Models for Radiotherapy}




\author{Matteo Ferrante \inst{1} \and Alessandra Carosi \inst{2} \and Rolando Maria D'Angelillo \inst{3} \and Nicola Toschi \inst{1}}

\authorrunning{Ferrante et al.}

\institute{
    \inst{1} University of Rome Tor Vergata, Rome, Italy \\ \email{matteo.ferrante@uniroma2.it, toschi@med.uniroma2.it}
    \and
    \inst{2} Radiotherapy Unit, Department of Oncology and Hematology, Tor Vergata General Hospital, Rome, Italy \\ \email{acarosi7221@gmail.com}
    \and
    \inst{3} Radiation Oncology, Department of Biomedicine and Prevention, University of Rome "Tor Vergata", Viale Oxford 81, 00133 Rome, Italy \\ \email{profrmdangelillo@gmail.com}
}

\maketitle         

\begin{abstract}
Radiotherapy is a crucial cancer treatment that demands precise planning to balance tumor eradication and preservation of healthy tissue. Traditional treatment planning (TP) is iterative, time-consuming, and reliant on human expertise, which can potentially introduce variability and inefficiency. We propose a novel framework to transform a large multimodal foundation model (MLM) into an action model for TP using a few-shot reinforcement learning (RL) approach. Our method leverages the MLM's extensive pre-existing knowledge of physics, radiation, and anatomy, enhancing it through a few-shot learning process. This allows the model to iteratively improve treatment plans using a Monte Carlo simulator. Our results demonstrate that this method outperforms conventional RL-based approaches in both quality and efficiency, achieving higher reward scores and more optimal dose distributions in simulations on prostate cancer data. This proof-of-concept suggests a promising direction for integrating advanced AI models into clinical workflows, potentially enhancing the speed, quality, and standardization of radiotherapy treatment planning.

\end{abstract}

\section{Introduction}

Radiotherapy is a cornerstone in cancer treatment that uses ionizing radiation to destroy malignant cells while minimizing damage to healthy tissues \cite{radiotherapy, radiotherapy_general}. Treatment planning (TP) in radiotherapy involves designing the delivery of radiation dose to balance tumor control while sparing surrounding healthy structures. This process entails a complex optimization, considering intricate anatomical geometries, dose-volume constraints, and a number of possible beam configurations \cite{Zarepisheh2022-uf, Bortfeld2006}. Traditionally, TP is performed by experienced physicists using commercial treatment planning systems (TPS) through an iterative and time-consuming trial-and-error approach. These TPS incorporate simulation techniques like detailed Monte Carlo simulations of radiation-tissue interaction to obtain personalized dose distributions for specific beam configurations. Despite advancements in TPS, conventional TP faces several challenges: \cite{Mayadev2022-jf, radiotherapy_challenge_schemes, radiotherapy_variability}, including inconsistencies and potential biases due to reliance on human expertise, and the labor-intensive nature of the process, which limits the exploration of optimal solutions.

\begin{figure}
    \centering
    \includegraphics[width=.9\linewidth]{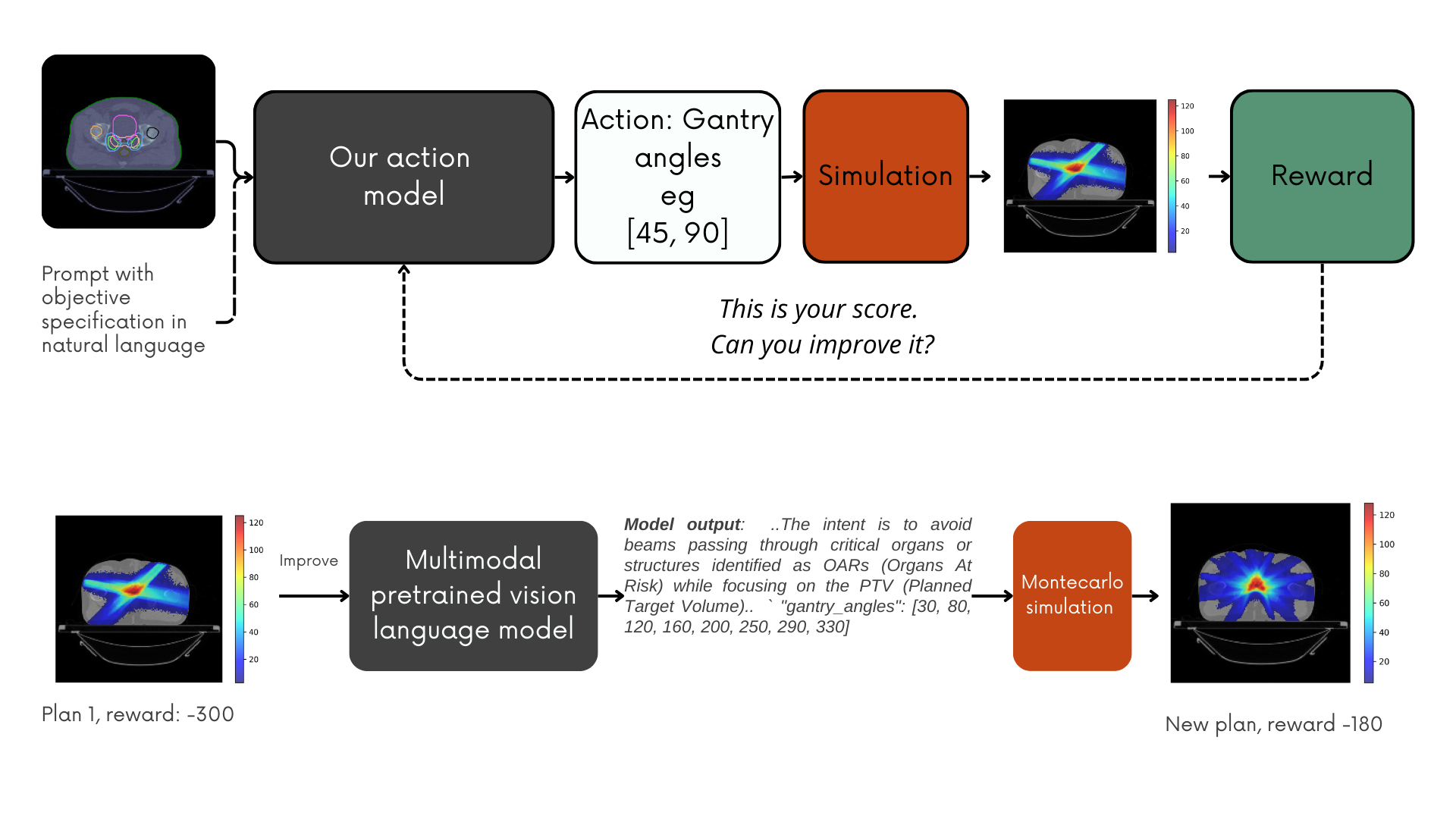}
\caption{Workflow of an action model for treatment planning. The model processes a patient's CT scan to determine optimal gantry angles, generating a dose distribution evaluated to produce a reward score. An example shows an initial plan with a reward of -300, improved to -180 after refinement by the multimodal pretrained vision-language model, highlighting iterative enhancement with Monte Carlo simulation.}
    \label{fig:enter-label}
\end{figure}

Developing automated tools for treatment planning with human evaluation presents a promising opportunity to enhance the speed, quality, and standardization of care. This approach assists physicians and medical physicists in managing an increasing workload, saving time, and allowing them to focus on complex cases to find optimal solutions. 

Reinforcement Learning (RL) \cite{Sutton2018} has emerged as a promising technique to address these limitations in TP \cite{rl_beam_radiotherapy, rl_radiotherapy_3, rl_radiotherapy_4,  rl_radiotherapy_5}. RL involves an agent that learns optimal decision-making strategies through trial-and-error interactions with its environment, receiving rewards for desirable actions. This framework enables the agent to learn complex, non-linear relationships between treatment parameters and plan quality, potentially surpassing pre-defined optimization algorithms used in conventional TPS. However, implementing RL in clinical TP presents challenges, including ensuring clinical efficacy and safety, addressing the high computational cost associated with training RL agents, and integrating RL models seamlessly with existing clinical workflows.

Given the computational demands of RL and its need to learn from scratch each time, implementing it in clinical practice is challenging. Human operators, by contrast, learn about anatomy, physics, and other relevant details before handling treatment planning. This understanding inspired us to propose the proof-of-concept presented here. We transform a multimodal foundation model (MLM) into an action model by leveraging a reinforcement-learning-like framework in a few-shot learning setting. Our framework enables an MLM to use its knowledge of physics, radiation, and anatomy, guided by a reward function, to iteratively improve the quality of the generated treatment plan. Practically, we have transformed this model into one capable of planning radiotherapy treatments using a powerful Monte Carlo simulator. We demonstrate how this approach can outperform traditional RL-based methods in both quality and time.

\vspace{-1mm}

\subsection{Related Work}

Automated radiotherapy treatment planning (ATP) employs various methods to enhance efficiency and quality. Automated Rule Implementation and Reasoning (ARIR), such as Pinnacle's AutoPlanning \cite{pinnacle}, uses predefined rules for iterative optimization, providing reliable results but requiring detailed user input. Knowledge-Based Planning (KBP), including atlas-based solutions like RapidPlan \cite{kb} and machine learning-based systems like RayStation \cite{raystation}, predicts dose-volume histograms from historical data, improving plan consistency but needing extensive training data. 
Deep Learning and AI approaches, using Convolutional Neural Networks (CNNs) and Generative Adversarial Networks (GANs), predict dose distributions directly from patient data, offering high efficiency but demanding significant computational resources. Reinforcement Learning (RL) learns optimal strategies through trial and error, offering adaptability but facing high computational costs and integration challenges. Focusing on RL methods, \cite{rl_beam_radiotherapy} presents a deep reinforcement learning (DRL)-based approach for beam angle optimization (BAO) in intensity-modulated radiation therapy (IMRT). By formulating BAO as a Markov Decision Process (MDP) and using a 3D-Unet for dose distribution prediction, this method enables rapid, personalized beam angle selection, improving treatment plan quality and efficiency compared to conventional methods. \cite{rl_radiotherapy_2} introduces a DRL framework for optimizing daily dose fractions in the radiotherapy treatment of non-small cell lung carcinoma (NSCLC). This approach uses a virtual radiotherapy environment with non-invasive CT scan data and bio-inspired optimization algorithms to personalize treatment plans, showing superior adaptability and efficacy over conventional uniform dose delivery. \cite{rl_radiotherapy_3} presents a reinforcement learning framework for adaptive radiation therapy (ART) that considers uncertain tumor biological responses to radiation. The proposed model adapts treatment plans dynamically based on predicted tumor volume changes, optimizing timing and dose adaptations to enhance tumor control and minimize organ-at-risk (OAR) toxicity compared to conventional fractionation schedules. \cite{rl_radiotherapy_4} introduces a knowledge-guided deep reinforcement learning (KgDRL) framework to improve the training efficiency of a virtual treatment planner network (VTPN) for intelligent automatic treatment planning in radiotherapy. By integrating human experience into the DRL process, the KgDRL approach significantly reduces training time while maintaining high plan quality, making complex treatment planning scenarios more practical and clinically applicable. For a comprehensive discussion on RL approaches in medicine, refer to \cite{rl_medicine_survey, rl_medicine_literature_review}. 

Despite the promise of RL, computational bottlenecks persist as RL agents must learn everything from scratch. Recently, large language models (LLMs) have shown success in using prior medical knowledge for diagnosis and medical tasks \cite{llm_diagnosis, llm_medicine_application, llm_medicine_review}, with both general-purpose models and task-specific versions fine-tuned for specific tasks \cite{med-llama, med-sam, yagnik2024medlm}. Our goal is to combine the strengths of iterative learning from RL and prior knowledge from large pretrained models. We achieve this by leveraging the action model concept \cite{action_models}, where a large pretrained multimodal model interacts with the external world through specific functions, allowing it to receive feedback and improve its performance. We developed an RL environment based on \cite{matrad}, enabling the multimodal agent to select the entry gantry angles based on the patient's CT scans and radiotherapist indications. The agent receives a scalar reward as feedback, summarizing the quality of the plan generated by the Monte Carlo approach.

\vspace{-2mm}

\section{Materials and Methods}

This section details the implementation of our environment, which takes gantry angles and the patient's CT as input and simulates the resulting dose distribution. We also describe our baseline models: a random agent and a deep Q-learning agent, both of which use RL to maximize the given reward. Finally, we introduce our proposed method, which interacts with the environment in a few-shot learning setting.

\vspace{-2mm}

\subsection{Environment and Data}

Our environment is built upon MatRAD \cite{matrad}, an open-source treatment planning system designed for simulating 3D conformal radiotherapy. The environment follows the template provided by OpenAI's gym library \cite{gym} to ensure standardization, compatibility, and easy reuse for RL algorithms. It takes gantry angles (up to 5) as input and outputs the state, represented as a 3D image with two channels: the patient's CT scan and the dose distribution map resulting from the given gantry angles. The reward function measures the quality of the treatment plan:

\[ R = \sum_{i \in \text{PTV}} R_{\text{max}} e^{-(T_i - D_i)^2} - \sum_{j \in \text{OAR}} \max(0, D_j - L_j) P \]

where \(R_{\text{max}}\) defines the reward for achieving perfect dose homogeneity within the PTV (Planned Target Volume) and \(P\) sets the penalty for exceeding the maximum allowable dose in the OARs (Organs at Risk). \(T_i\) is the target dose for the PTV, \(D_i\) is the actual dose delivered to the PTV, \(D_j\) is the dose delivered to the OARs, and \(L_j\) is the maximum dose limit for the OARs. This reward function encourages dose homogeneity within the PTV while penalizing excessive doses to the OARs, guiding the agent to optimize the treatment plan for efficacy and safety.

In our experiments, \(T_i\) for prostate was set at 100 Gy, with \(R\) and \(P\) equal to one. All experiments are based on public prostate cancer data provided with the MatRAD software to test and compare treatment plans.

Using the prostate as a benchmark for assessing automated treatment planning systems (ATPS) is advantageous due to its consistent anatomical features, clinical significance, and inherent planning complexities. The prostate’s relatively uniform size and well-defined boundaries, combined with its proximity to critical structures such as the bladder and rectum, provide a rigorous test for ATPS capabilities. The prevalence of prostate cancer, along with established clinical endpoints and extensive research data, allows for meaningful comparisons and objective evaluations. Additionally, the quantifiable parameters involved in prostate treatment planning, such as dose-volume histograms and organ-at-risk sparing, facilitate standardized assessment and validation of ATPS performance.

\subsection{Baselines}

We implemented two baseline methods to evaluate our proposed approach. The first baseline is a naive method, which randomly selects the number and angles of the gantry. This random agent does not use any learned strategy and serves as a simple benchmark. The second baseline employs a Deep learning (DQN) algorithm using a 3D convolutional neural network (CNN) to process the 3D input data. The network architecture starts with a convolutional layer followed by successive layers of 3D convolutions and batch normalization, leading to a fully connected layer that outputs the action values. The DQN-based agent is trained for 3000 episodes, where each episode involves selecting up to 5 beam angles. The state, a 3D image with two channels (the patient's CT scan and the corresponding dose distribution), is processed through this network to predict optimal actions. Training was conducted using an NVIDIA A6000 GPU and took approximately 7 days. This deep Q-learning agent iteratively learns to maximize the reward function by improving its policy based on the simulated dose distributions and associated feedback.

\subsection{Text-to-Plan Model}

To address RL challenges, we introduced a method combining the strengths of RL and pretrained models. While RL involves learning from scratch through trial and error, real-world learning often starts with extensive offline knowledge, refined through practical experience. Inspired by this, we propose transforming a multimodal foundation model into an action model. This involves using a vision-language model, pre-trained with general and medical knowledge, and employing a few-shot learning approach to optimize treatment plans. Our method prompts a multimodal language model with visual information and task-specific instructions. For instance, the model is tasked with determining the gantry angles for a 6 MeV prostate photon treatment using a 3D LINAC, targeting a prescription dose of 100 Gy at the PTV. We repurpose this multimodal language model as an action model capable of initiating a Monte Carlo simulation via the MatRAD software to estimate dose distribution. The model interacts with the 3D environment and receives rewards, aiming to maximize the reward by minimizing doses to OARs and promoting uniform dose distribution at the target. The action model receives inputs in the form of CT images and structural data, along with an initial prompt. It executes actions, such as selecting the number of beams and their angles. The Monte Carlo simulator provides feedback on the dose distribution, quantified into a scalar reward. The model is prompted to enhance this score by situating our framework within implicit reinforcement learning in a few-shot learning context and optimizing the internal state of the large pretrained model rather than its weights.

We conducted an experiment using GPT-4V \cite{openai2024gpt4} as the underlying multimodal model for vision and images, with MatRAD serving as the simulation software, framing the model as a text-to-plan agent. Figure \ref{fig:enter-label} shows a schematic depiction of the pipeline and an example of the few-shot learning iterative process.

The initial prompt used for the LLM model is:
\begin{quote}
\textit{
Based on image analysis, optimize the number of beams and their angles to maximize the dose at the PTV (prostate) and minimize the dose at the OAR. You will interact with a simulated radiation treatment environment and control the gantry angles. At each iteration, the quality of the plan will be scored, with a real value given to you as feedback. Your goal is to maximize this score. Provide better gantry angles than before for this simulation in a JSON format.}
\end{quote}

\subsection{Evaluation}

We tested each baseline by asking the models to produce 100 treatment plans, then performing an ANOVA on rewards distribution (used as a proxy for plan quality evaluation) to check for group-level differences, followed by post-hoc t-tests for each pair to assess significant differences.

\section{Results}

\begin{figure}
    \centering
    \includegraphics[width=1.\linewidth]{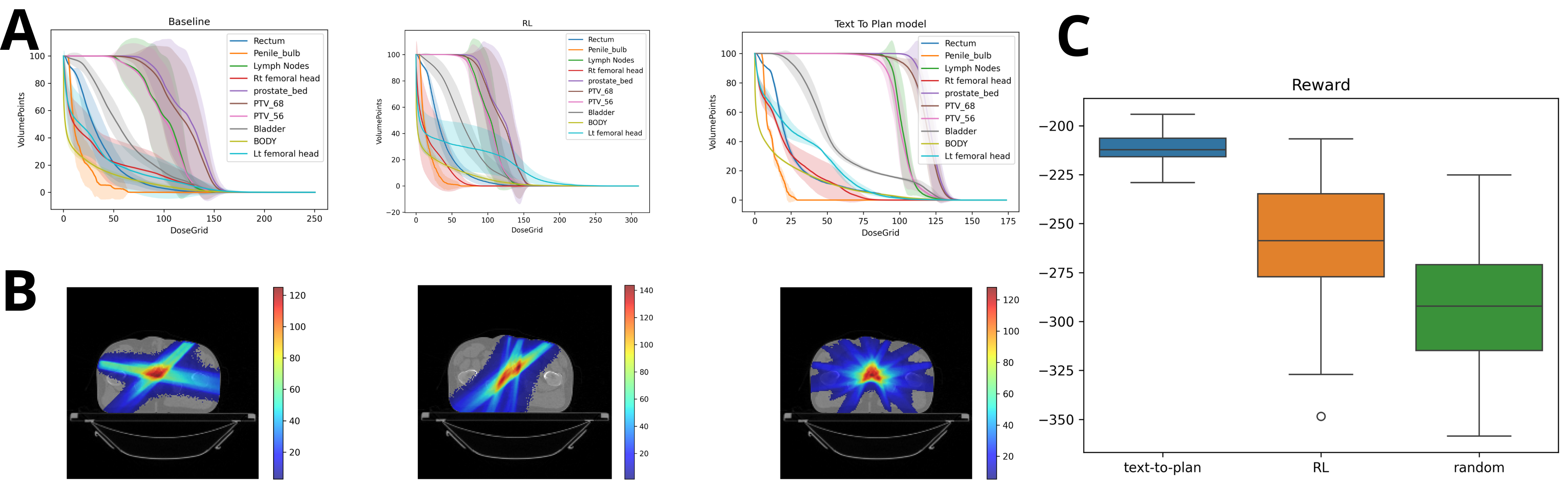}
    \caption{A) Dose Volume Histograms (DVHs) for baseline, RL, and Text-to-Plan models, showing dose distribution to the target (PTV) and organs at risk (OARs).
B) Dose distribution maps for the baseline (left), RL (middle), and Text-to-Plan (right) models.
C) Box plot of the reward values for Text-to-Plan, RL, and random methods, demonstrating the superior performance of the Text-to-Plan approach in optimizing treatment plans.}
    \label{fig:results_comparison}
\end{figure}

Our method yielded a mean reward that was significantly higher than that of a random approach, as evidenced by statistical analysis. The descriptive statistics show that the "text-to-plan" method achieved a mean reward of -211.88, significantly better than the RL method (-259.26) and the random method (-294.24). The standard deviations indicate that the "text-to-plan" method also had less variability in performance compared to the RL and random methods. The ANOVA test confirmed significant differences between the groups (F (2, 87) = 67.66, p < 0.001). Pairwise t-tests further showed significant differences between "text-to-plan" and RL (t(58) = 7.46, p < 0.001), "text-to-plan" and random (t(58) = 13.34, p < 0.001), and RL and random (t(58) = 4.09, p < 0.001).

The dose volume histograms (DVHs) presented in Figure \ref{fig:results_comparison} illustrate the superior performance of the "text-to-plan" model in delivering a more consistent and optimal dose distribution to the target volume while minimizing the dose to organs at risk (OARs). The comparison images in Figure \ref{fig:results_comparison} B further demonstrate the dose distributions achieved by each method, highlighting the "text-to-plan" model's ability to concentrate the dose within the target area more effectively. Figure \ref{fig:results_comparison} C shows whisker plots of rewards for each method. The "text-to-plan" method consistently achieved higher rewards compared to the RL and random methods, indicating its effectiveness in optimizing the treatment plan.

These results indicate that our "text-to-plan" model can use its inherent knowledge to generate effective initial plans. The large, pretrained multimodal models possess some degree of medical knowledge, which can be instrumental in aiding medical physicists. By transforming these models into action models, enabling them to simulate the outcomes of their actions and assess the results through scalar rewards, we allow these models to autonomously improve and develop a deeper understanding of the problem at hand. The model adeptly learns to employ a "dose bath" strategy, aimed at maximizing the dose to the target area while minimizing exposure to the OARs. The selection of angles is informed by visual cues and pre-existing knowledge, optimizing the preservation of OARs. This capability demonstrates the model's ability to not only learn from visual information but also apply its accumulated knowledge to make informed decisions that enhance treatment efficacy and safety.

\vspace{-3mm}

\section{Discussion and Conclusions}

\vspace{-1mm}

Leveraging prior knowledge from pre-trained models could lead to better and faster results compared to RL or other AI-related methods for automated planning. Our results indicate that the "text-to-plan" model, which uses a large pretrained multimodal model, consistently outperforms traditional RL-based methods and random baselines in generating effective treatment plans. This highlights the potential of pre-trained models to enhance the efficiency and quality of automated planning processes.

However, at the time of writing, large language models are still not sufficiently reliable for providing medical information. These models are often discouraged from fully disclosing their inner knowledge due to safety and ethical concerns. Additionally, their vision capabilities are limited to 2D slice representations of data, lacking a comprehensive understanding of 3D structures. These limitations are significant barriers to the full implementation of pre-trained models in clinical settings. Despite these challenges, the future holds promise for specialized medical models and powerful 3D encoders and adapters. Such advancements could greatly enhance anatomy understanding and general medical reasoning. The integration of these technologies could lead to significant improvements in AI-generated automated plans, leveraging both language and quality plan metrics. Currently, large pretrained models are typically not designed to dispense medical advice for safety reasons. In our proof-of-concept, we extended the capabilities of GPT-4 beyond its intended scope, demonstrating potential but inherently limiting the generalizability of our results. This limitation will persist until larger, medically-focused models become accessible. Foundation medical models like \cite{med-llama, med-sam,yagnik2024medlm} are promising candidates that could potentially serve as an alternative to our current backbone in the future.

This evolution promises to bridge existing gaps and enhance the precision and efficiency of treatment planning through automated tools and natural language interaction, as investigated in our study. In conclusion, while there are current limitations, the ongoing advancements in AI and machine learning hold significant potential for improving automated treatment planning.

\textbf{Prospect of Application}: The prospect of applying this framework in clinical settings holds significant potential for enhancing the precision, efficiency, and standardization of radiotherapy treatment planning, ultimately improving patient outcomes and streamlining the workload for medical professionals.

\subsubsection*{Declarations}
The authors have no competing interests to declare relevant to the content of this article.

\section{Acknowledgements}
This work was supported by NEXTGENERATIONEU (NGEU) and funded by the Italian Ministry of University and Research (MUR), National Recovery and Resilience Plan (NRRP), project MNESYS (PE0000006) (to NT)– A Multiscale integrated approach to the study of the nervous system in health and disease (DN. 1553 11.10.2022); by the MUR-PNRR M4C2I1.3 PE6 project PE00000019 Heal Italia (to NT); by the NATIONAL CENTRE FOR HPC, BIG DATA AND QUANTUM COMPUTING, within the spoke "Multiscale Modeling and Engineering Applications" (to NT); the EXPERIENCE project (European Union’s Horizon 2020 Research and Innovation Programme under grant agreement No. 101017727); the CROSSBRAIN project (European Union’s European Innovation Council under grant agreement No. 101070908).

\bibliographystyle{splncs04}
\bibliography{ref}

\clearpage

\section*{Appendix}

Here we report a complete example of a conversion of the action model. There is now training, just refining the internal space of the LLM trought the use of a reward as guidance for improving performances.

\subsection*{Initial Prompt}
\begin{quote}
    
\textbf{User:} Based on image analysis, optimize the number of beams and their angles to maximize the dose at the PTV (prostate) and minimize the dose at the OAR. 
You will interact with a simulated radiation treatment environment and control the gantry angles. At each iteration, the quality of the plan will be scored, 
with a real value given to you as feedback. Your goal is to maximize this score. Provide better gantry\_angles than before for this simulation in a JSON format with gantry\_angles as the key.
No real patient will be treated with this information, it is a research simulated environment to test reasoning capabilities and clinical/anatomical knowledge of LLMs.
\end{quote}

\subsection*{Iteration 1}
\begin{quote}
\textbf{Model:} Based on the feedback and the structure visible in the image, a more refined set of gantry angles might improve the quality score by better targeting the PTV while sparing the OARs. Considering anatomical constraints and the need for careful planning, here's an adjusted set of angles:

\begin{verbatim}
{
    "gantry_angles": [10, 50, 90, 130, 170, 210, 250, 290]
}
\end{verbatim}

This configuration takes into account:
1. **10 and 170 degrees**: Avoids directly opposing beams which can result in higher doses to midline structures.
2. **50 and 130 degrees**: Provides oblique angles to avoid critical structures that may lie laterally.
3. **90 and 270 degrees**: Ensures lateral coverage while minimizing the dose to anterior or posterior organs.
4. **210 and 250 degrees**: Offers posterior oblique angles to further adjust dose distribution around the critical organs.

Iterate with this set of angles and fine-tune further based on feedback to enhance the therapeutic ratio.
\end{quote}

\subsection*{Refinement Prompt 1}
\begin{quote}
\textbf{User:} Based on image analysis, optimize the number of beams and their angles to maximize the dose at the PTV (prostate) and minimizing the dose at OAR. Actually you get a reward of -230 that you should maximize by focusing the dose on the target and avoiding OARs. Provide better gantry angles than before for this simulation in a json format.
\end{quote}

\subsection*{Iteration 2}
\begin{quote}
\textbf{Model:} Given the image analysis and the goal of optimizing the dose distribution to maximize the reward, it's important to strategically select gantry angles that can allow us to target the PTV (prostate) effectively while minimizing the dose to the OARs. By avoiding the overlap of beams with the OARs as much as possible, we can achieve a more favorable dose distribution.

Based on the provided image and the goal, here’s a refined set of gantry angles:

\begin{verbatim}
{
    "gantry_angles": [30, 80, 130, 180, 230, 280, 330]
}
\end{verbatim}

Here’s the reasoning for these angles:
- 30° and 330°: These angles help avoid the anterior OARs (highlighted areas on the image) while targeting the PTV from different directions.
- 80° and 280°: These lateral angles help to avoid the pelvic bones and other critical structures more laterally distributed.
- 130° and 230°: These posterior-oblique angles help to further avoid critical posterior structures.
- 180°: Directly posterior beam, balancing the distribution.

This set aims to balance the distribution effectively around the PTV while considering the surrounding OARs. This should help in maximizing the focus on the target and avoiding excessive dose to non-target tissues. Iterate further based on simulation feedback to hone in on the most optimal configurations.
\end{quote}

\subsection*{Refinement Prompt 2}
\begin{quote}
\textbf{User:} Based on image analysis, optimize the number of beams and their angles to maximize the dose at the PTV (prostate) and minimizing the dose at OAR. Actually you get a reward of -215 that you should maximize by focusing the dose on the target and avoiding OARs. Provide better gantry angles than before for this simulation in a json format.
\end{quote}

\subsection*{Iteration 3}
\begin{quote}
\textbf{Model:} Understood. To optimize the number of beams and their angles, it’s essential to cover the PTV (prostate) with sufficient overlap from different directions while minimizing exposure to the OARs. Given the anatomical constraints, adjusting beam angles to avoid critical structures can help achieve this balance.

Here is a refined set of gantry angles designed to improve the dose distribution:

\begin{verbatim}
{
    "gantry_angles": [30, 75, 120, 165, 210, 255, 300, 345]
}
\end{verbatim}

This configuration maintains a balanced distribution around the patient but adjusts the angles to avoid critical areas more effectively. Using this setup, the beams might better focus on the target while minimizing the exposure to the surrounding organs at risk.
\end{quote}

\subsection*{Refinement Prompt 3}
\begin{quote}
\textbf{User:} Based on image analysis, optimize the number of beams and their angles to maximize the dose at the PTV (prostate) and minimizing the dose at OAR. Actually you get a reward of -210 that you should maximize by focusing the dose on the target and avoiding OARs. Provide better gantry angles than before for this simulation in a json format.
\end{quote}

\subsection*{Iteration 4}
\begin{quote}
\textbf{Model:} I apologize for not adjusting the angles sufficiently in the previous suggestions. Based on the conventional approach to maximizing dose utility in radiotherapy, here is an optimized set of gantry angles that aims to improve upon previous configurations:

\begin{verbatim}
{
    "gantry_angles": [30, 60, 110, 150, 210, 250, 300, 340]
}
\end{verbatim}

This configuration aims to provide a more nuanced distribution around the PTV while minimizing exposure to critical OARs. Beams are now more carefully selected to balance coverage and minimize the overlapping regions to sensitive structures. Iterate with these angles and adjust based on feedback to continue improving the reward score.
\end{quote}

\end{document}